\pdfoutput=1

\documentclass[11pt]{article}

\usepackage[preprint]{acl}

\usepackage{times}
\usepackage{latexsym}

\usepackage[T1]{fontenc}

\usepackage[utf8]{inputenc}

\usepackage{microtype}

\usepackage{inconsolata}

\usepackage{subcaption}
\usepackage{multirow}
\usepackage{booktabs}

\usepackage{graphicx}
\usepackage{colortbl}
\usepackage{latexsym}
\usepackage{array}  
\usepackage{inconsolata}
\usepackage{mathtools}
\usepackage{extarrows}
\usepackage{amsmath}
\usepackage{arydshln}
\usepackage{array}
\usepackage{amssymb}
\usepackage{booktabs}
\usepackage{amsfonts}
\usepackage{mathtools,amssymb}
\usepackage{bbm, dsfont}
\usepackage{bm}
\usepackage{bbm}
\usepackage[ruled,vlined,linesnumbered]{algorithm2e}
\usepackage{enumitem}

\definecolor{ourscolor}{RGB}{220,240,255}
\usepackage{booktabs} %
\usepackage{array}    %
\usepackage{graphicx} %
\usepackage{amsmath}  %
\usepackage{textcomp} %

\title{\textsc{HS-STaR}: Hierarchical Sampling for Self-Taught Reasoners via\\Difficulty Estimation and Budget Reallocation}

\author{
 \textbf{Feng Xiong\textsuperscript{1}}\footnotemark[1],
 \textbf{Hongling Xu\footnotemark[1]},
 \textbf{Yifei Wang\textsuperscript{1,2}},
 \textbf{Runxi Cheng\textsuperscript{3}},
 \textbf{Yong Wang\textsuperscript{1}}\footnotemark[2],
 \textbf{Xiangxiang Chu\textsuperscript{1}}
\\
 \textsuperscript{1}AMAP, Alibaba Group \textsuperscript{2}University of Chinese Academy of Sciences \textsuperscript{3}Tsinghua University
\\
\texttt{jingxun.xf@icloud.com, wangyong.lz@alibaba-inc.com} 
}

\begin{document}
\maketitle
{
\renewcommand{\thefootnote}{\fnsymbol{footnote}}
\footnotetext[1]{\ Equal contribution.}
\footnotetext[2]{\ Corresponding author and project lead.}
}

\begin{abstract}
Self-taught reasoners (STaRs) enhance the mathematical reasoning abilities of large language models (LLMs) by leveraging self-generated responses for self-training. Recent studies have incorporated reward models to guide response selection or decoding, aiming to obtain higher-quality data. However, they typically allocate a uniform sampling budget across all problems, overlooking the varying utility of problems at different difficulty levels. In this work, we conduct an empirical study and find that problems near the boundary of the LLM's reasoning capability offer significantly greater learning utility than both easy and overly difficult ones. To identify and exploit such problems, we propose \textsc{HS-STaR}, a \textbf{H}ierarchical \textbf{S}ampling framework for \textbf{S}elf-\textbf{Ta}ught \textbf{R}easoners. Given a fixed sampling budget, \textsc{HS-STaR} first performs lightweight pre-sampling with a reward-guided difficulty estimation strategy to efficiently identify boundary-level problems. Subsequently, it dynamically reallocates the remaining budget toward these high-utility problems during a re-sampling phase, maximizing the generation of valuable training data. Extensive experiments across multiple reasoning benchmarks and backbone LLMs demonstrate that \textsc{HS-STaR} significantly outperforms other baselines without requiring additional sampling budget.

\end{abstract}

\section{Introduction}

Large language models (LLMs) can improve their capabilities by training on self-generated data, characterizing them as self-taught reasoners (STaRs)~\cite{zelikman2022star,yuan2023_rejection_sampling_finetune,hosseini2024vstar}. This paradigm is also referred to as reinforced self-training~\cite{gulcehre2023_rest} (ReST) or self-improvement~\cite{huang-etal-2023-lmsi}.
For mathematical reasoning~\cite{yang2025llmreasoningdistractedirrelevant, tian2025vcsearchbridginggap, wang2025code}, pioneer STaRs generally follow an \textbf{iterative process}: (1) generating candidate responses for a given math problem via temperature sampling; (2) selecting responses based on answer correctness; and (3) updating the model using either SFT or DPO~\cite{singh2024beyond, pang2024iterative, wu2025_progress_or_regress}.

Building on previous efforts, recent work has focused on enhancing STaRs by leveraging additional reward models, which can be categorized into two main directions. 
One line of work, known as \textit{reward-guided selection}, introduces an auxiliary reward model to re-rank or filter responses based on their estimated quality, encouraging the model to exploit higher-quality trajectories~\cite{yang2024qwen25mathtechnicalreportmathematical,zeng2025bstar,tu2025enhancingllmreasoningiterative}.  
Another line of work, \textit{reward-guided decoding}, leverages Monte Carlo Tree Search (MCTS), in which a process reward model (PRM)~\cite{wang-etal-2024-math} is trained and used to guide the decoding process, aiming to improve both final answer accuracy and the quality of intermediate reasoning steps~\cite{zhang2024restmcts, chen2024alphamath}.

\begin{figure*}[ht]
\centering
\includegraphics[width=1\linewidth]{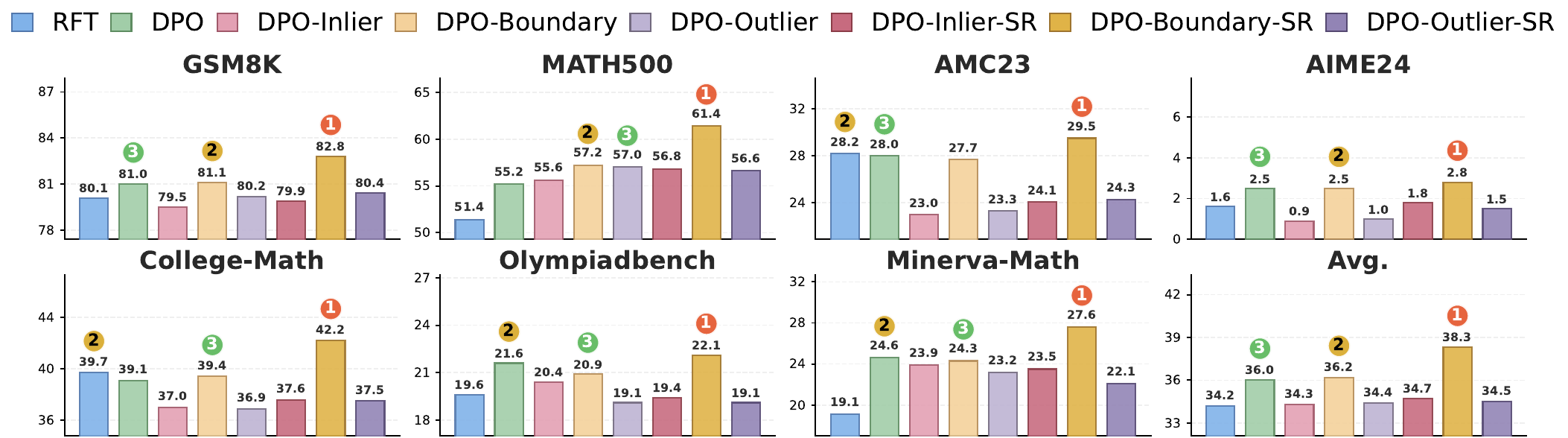}
\caption{
Pilot experiments on Qwen2.5-3B. 
}
\label{fig:intro_benchmark_compact}
\end{figure*}

However, these studies primarily focus on response quality through such reward-guided approaches, neglecting the utility of the problems themselves. 
Specifically, a uniform allocation of sampling budget across all problems fails to account for the varying difficulty levels of individual problems and their differential impacts on the learning process~\cite{zhu2024_exploring_learning_difficulty_of_data}.
Since auto-regressive decoding is the principal bottleneck in STaRs, such an indiscriminate allocation strategy is highly inefficient.
This issue raises two critical questions: 
\textit{(i)} \textit{Profiling:} \textit{Which difficulty level of problems are most beneficial for self-taught reasoners?} Intuitively, problems that are too simple provide limited learning value, while those that are overly challenging may either waste sampling resources by requiring numerous attempts to generate correct responses or be beyond the model’s capabilities, hindering effective learning. \textit{(ii) Allocation: How can sampling resources be allocated to maximize the utility of valuable problems?} Given the high expense of sampling, it is essential to identify and prioritize high-utility problems to optimize the trade-off between resource usage and performance improvement.

To address these questions, we first conduct a pilot study to analyze the utility of problems across varying difficulty levels (see Sec.~\ref{sec:pilot} for details).  
We begin by defining model-specific problem difficulty based on the accuracy over multiple sampling attempts~\cite{snell2025scaling, tong_dart_math}.  
As depicted in Fig.~\ref{fig:intro_benchmark_compact}, we observe that training solely on either \textit{Inlier} or \textit{Outlier} problems leads to a significant decline in performance, whereas training exclusively on \textit{Boundary} problems yields even better results than using the full set of problems.
Furthermore, allocating additional sampling budget to these \textit{Boundary} problems for self-training substantially improves model performance, which underscores their importance in guiding more effective learning in STaRs.

While the above findings highlight the high utility of \textit{Boundary} problems, identifying them typically relies on statistical estimation with extensive sampling, limiting the practical applicability. To address this limitation and further tackle the second question, we propose the \textbf{H}ierarchical \textbf{S}ampling framework for \textbf{S}elf-\textbf{Ta}ught \textbf{R}easoners (\textsc{HS-STaR}). 
Given a fixed total sampling budget, \textsc{HS-STaR} begins with a \textit{Difficulty Estimation} phase, where a small portion of the budget is used to estimate problem difficulty based on both answer accuracy and response quality, a process we refer to as reward-guided difficulty estimation. The remaining budget is then dynamically reallocated to problems estimated to be of high utility in a subsequent \textit{Re-Sampling} phase, thereby maximizing the exploitation of valuable problems without incurring additional budget. Finally, the aggregated responses are used to construct a preference dataset for self-training in \textit{Preference Optimization} phase, improving the overall effectiveness of STaRs.

Our contributions are summarized as follows:
\begin{itemize}[itemsep=0.5pt, topsep=1.5pt]
    \item We conduct an empirical study that reveals the high utility of \textit{Boundary} problems in self-taught reasoning. This motivates a problem-centric perspective for optimizing sampling resource allocation by identifying and prioritizing these problems.

    \item We propose \textsc{HS-STaR}, a hierarchical sampling framework that integrates reward-guided difficulty estimation to dynamically reallocate sampling budgets toward high-utility problems, significantly enhancing training effectiveness under a fixed sampling budget.

    \item Extensive experiments across seven reasoning benchmarks and various backbone LLMs demonstrate the superiority of our \textsc{HS-STaR}. Further analyses confirm the effectiveness of each component within the framework.
    
\end{itemize}

\section{Pilot Experiments}
\label{sec:pilot}

To analyze the core challenges of \textit{Profiling} and \textit{Allocation}, we conduct a comprehensive empirical study on the utility of problems in STaRs.

\subsection{Preliminary}
\label{sec: preliminary}

We begin by formalizing the iterative self-training process of STaRs. 
At iteration $t$, we denote the policy model as $\mathcal{M}_t$ and the utilized dataset as $\mathcal{D}_t = \{(x_i, y_i)\}_{i=1}^N$, where $x_i$ is a math problem and $y_i$ is the corresponding answer. This process typically consists of three steps:

\noindent\textbf{(1) Generation.}
For each problem $x \in \mathcal{D}_t$, the model $\mathcal{M}_t$ generates $n$ responses by sampling, forming the set $\mathcal{R}_{t,x} = \{r_{j}|r_{j}\sim\mathcal{M}_t(x) \}_{j=1}^{n}$.

\noindent\textbf{(2) Selection.}
We apply a rule-based verifier $V(x, y, r) \in \{0, 1\}$ to assess response correctness, and omit the answer $y$ from the notation hereafter for simplicity.
For Rejection Sampling Fine-Tuning~(RFT)~\cite{yuan2023_rejection_sampling_finetune}, we select only correct responses to form $\mathcal{D}_t^{\text{corr}}$.  
For DPO~\cite{rafailov2023direct}, we construct $\mathcal{D}_t^{\text{pairs}}$ by pairing correct and incorrect responses.

\noindent\textbf{(3) Updating.}
For RFT, the model $\mathcal{M}$ is updated by minimizing the negative log-likelihood:
\begin{align}
\mathcal{L}_{\text{RFT}} = - \log \mathcal{M}(r|x),
\end{align}
where $(x,r)\in\mathcal{D}_t^{\text{corr}}$. For DPO, the model $\mathcal{M}$ is updated by minimizing: 
\begin{align}
\label{dpo_loss}
\mathcal{L}_{\text{DPO}}\hspace{-1mm} =\hspace{-1mm}-\hspace{-1mm}\log \sigma\hspace{-1mm}\left( 
\hspace{-1mm}\beta \left( \hspace{-1mm}
\log\hspace{-1mm} \frac{\mathcal{M}(r_w | x)}{\mathcal{M}_{\text{t}}(r_w | x)} 
\hspace{-1mm}
-
\hspace{-1mm} 
\log \frac{\mathcal{M}(r_l | x)}{\mathcal{M}_{\text{t}}(r_l | x)}
\hspace{-1mm} 
\right) 
\hspace{-1mm}
\right),\hspace{-1mm}
\end{align}
where $(x, r_w, r_l)\in\mathcal{D}_t^{\text{pairs}}$, $r_w$ is a correct response and $r_l$ is an incorrect response for problem $x$.

Additionally, we introduce Statistical Difficulty Estimation (\textbf{SDE}) as an oracle for assessing problem difficulty.  
Following~\citet{snell2025scaling}, SDE computes accuracy using a substantial sampling budget (i.e., 100 samples per problem), providing a reliable proxy for the model-specific difficulty of each problem. 
Inspired by~\citet{chen2024unlocking}, we partition problem instances as \emph{Inlier} (accuracy $>$ 87.5\%), \emph{Outlier} (accuracy $<$ 12.5\%), or \emph{Boundary} (otherwise), with their corresponding sets denoted as $\mathcal{D}_t^{\mathcal{I}}$, $\mathcal{D}_t^{\mathcal{O}}$, and $\mathcal{D}_t^{\mathcal{B}}$, respectively.

\begin{figure*}[t]
\centering
\includegraphics[width=\linewidth]{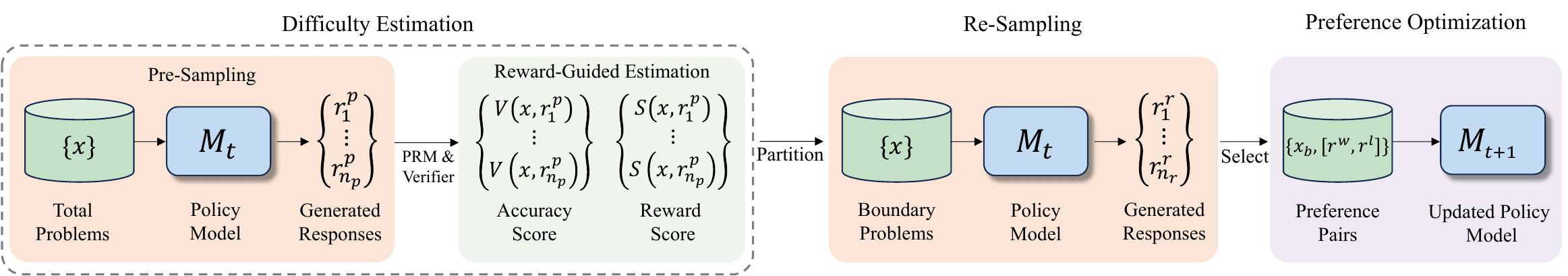}
\caption{
Illustration of the \textsc{HS-STaR} framework.  
Each iteration begins with a \textit{Difficulty Estimation} phase, where a limited sampling budget is used to generate candidate responses for each query, referred to as pre-sampling.  
These responses are then evaluated using a reward-guided strategy to estimate problem difficulty.  
In the subsequent \textit{Re-Sampling} phase, the remaining budget is allocated to high-utility boundary problems identified in the previous step. 
Finally, in the \textit{Preference Optimization} phase, preference pairs are constructed from all collected responses and used to update the policy model.
}

\label{fig:method}
\end{figure*}

\subsection{Analysis of Utility over Problem Difficulty}

We conduct experiments of STaRs using our SDE-based partitioning on Qwen-2.5 3B~\cite{qwen2025qwen25technicalreport}, as shown in Fig.~\ref{fig:intro_benchmark_compact}. Training and dataset details are provided in Sec.~\ref{sec: setup}.

\paragraph{Difficulty-Aware Training Analysis.}
Given the observed superiority of DPO over RFT, we adopt DPO as the default training objective throughout our study.  
We define three variants—\texttt{DPO-Inlier}, \texttt{DPO-Boundary}, and \texttt{DPO-Outlier}—each trained exclusively on one SDE-defined subset: $\mathcal{D}^{\mathcal{I}}$, $\mathcal{D}^{\mathcal{B}}$, and $\mathcal{D}^{\mathcal{O}}$, respectively, using a fixed sampling count of $n$ as 8.  
As shown in Fig.~\ref{fig:intro_benchmark_compact}, we find that both \texttt{DPO-Inlier} and \texttt{DPO-Outlier} yield significantly worse performance, while \texttt{DPO-Boundary} produces a slight improvement (+0.2\%) than using all problems.
These results highlight that boundary-level problems offer the highest utility for STaRs.

\paragraph{Sampling Budget Reallocation.}  
We further examine whether allocating more sampling resources to different difficulty levels can enhance self-training effectiveness.  
We introduce three Sampling Reallocation (SR) variants: \texttt{DPO-Inlier-SR}, \texttt{DPO-Boundary-SR}, and \texttt{DPO-Outlier-SR}, where the total sampling budget ($8 \times |\mathcal{D}_t|$) is reallocated exclusively to one difficulty category.  
This concentrated sampling allows each selected problem to receive more candidate responses. 
Notably, \texttt{DPO-Boundary-SR} significantly achieves the best performance across all benchmarks, with an average score of 38.3\%.  
These results reinforce that boundary-level problems are more sampling-efficient, indicating strategically prioritizing them is key to enhance self-training.

\section{Methodology}

In this section, we provide a detailed introduction to our \textsc{HS-STaR}, as shown in Fig.~\ref{fig:method}.
Our approach is divided into three main phases: Difficulty Estimation, Re-Sampling, and Preference Optimization.

\subsection{\textit{Phase 1}: Difficulty Estimation}

While SDE provides a reliable oracle for assessing problem difficulty, it requires extensive sampling and is computationally expensive.
To enable practical difficulty estimation under limited resources, we propose a lightweight alternative, including pre-sampling and reward-guided estimation.

\noindent\textbf{Pre-Sampling.}
We first perform a pre-sampling step, where a small portion of the sampling budget is used to generate responses for each question.  Concretely, given a problem $x\in\mathcal{D}_t$, we derive $n_p$ responses from the policy model $\mathcal{M}_t$, forming $\mathcal{R}_{t,x}^p = \{r_{1}, \ldots, r_{n_p} \mid r_i \sim \mathcal{M}_t(x) \}$. Here, $n_p$ is set to a relatively small value, allowing more remaining budget to be reallocated toward high-utility problems in \textit{Phase 2}. 

\noindent\textbf{Reward-Guided Estimation.}
Subsequently, we evaluate $\mathcal{R}_{t,x}^p$ using both the ground-truth answer and process reward model (PRM). Specifically, we propose a reward-guided difficulty estimation~(RDE) strategy, which incorporates two complementary metrics:  $\phi_{\mathrm{a}}(\mathcal{R}_{t,x}^p)$ for assessing accuracy, and $\phi_{\mathrm{r}}({\mathcal{R}}_{t,x}^p)$ for evaluating the quality of the underlying reasoning process.
$\phi_{\mathrm{a}}(\mathcal{R}_{t,x}^p)$ is defined as the average accuracy over all generated responses: $\phi_{\mathrm{a}}(\mathcal{R}_{t,x}^p) = \frac{1}{n_p} \sum_{i=1}^{n_p} V(x,r_i)$.
The term $\phi_{\mathrm{r}}({\mathcal{R}}_{t,x}^p)$ assesses the quality of the reasoning process produced by the policy model, which is achieved by leveraging the scores provided by the PRM for each sampled response. We define an aggregate process quality score as the average of the reward values within all sampled responses: $\phi_{\mathrm{r}}(\mathcal{R}_{t,x}^p) = \frac{1}{n_p}\sum_{i=1}^{n_p}S\left(r_{i}\right)$,
where $S(r_{i})$ represents the reward score assigned to the $i$-th response $r_{i}$. 
Given that a response $r_i$ consists of $n_i$ reasoning steps, with step $j$ assigned a reward score $s_{i,j}$, the overall process quality score for a complete response $r_i$ is defined as the minimum reward score across all steps in the sequence~\cite{tu2025enhancingllmreasoningiterative}:
$S\left(r_{i}\right)= \min_{j \in \{1, 2, \ldots, n_i\}} \{ s_{i,j} \}$.

Based on these two critical dimensions of responses, we categorize the difficulty level for the given model $\mathcal{M}_t$ on a specific problem $x$ into three distinct classes:
\begin{align}
\Phi_{\mathcal{M}_t}(x)\hspace{-1mm}=\hspace{-1.5mm}
\begin{cases}
\mathrm{\textit{Inlier}},\hspace{2.6mm}\text{if } \phi_{\mathrm{a}}(\mathcal{R}_{t,x}^p) \hspace{-0.6mm}=\hspace{-0.6mm} 1\hspace{-0.6mm} \land\hspace{-0.6mm} \phi_{\mathrm{r}}(\mathcal{R}_{t,x}^p) \hspace{-0.6mm}>\hspace{-0.6mm} \tau_h \\
\mathrm{\textit{Outlier}}, \text{if } \phi_{\mathrm{a}}(\mathcal{R}_{t,x}^p) \hspace{-0.6mm}=\hspace{-0.6mm} 0\hspace{-0.6mm} \land \phi_{\mathrm{r}}(\mathcal{R}_{t,x}^p) \hspace{-0.6mm}< \hspace{-0.6mm}\tau_l \\
\mathrm{\textit{Boundary}},  \text{otherwise}
\end{cases}\hspace{-0.2mm},
\end{align}
where $\tau_h$ and $\tau_l$ are predefined thresholds. This metric jointly captures the model's ability to solve a given problem by evaluating both the accuracy of the final answer and the soundness of the reasoning process, thereby providing an effective estimate of problem difficulty even with limited responses.

\subsection{\textit{Phase 2}: Re-Sampling}

Building on the insights from Sec.~\ref{sec:pilot}, which highlight the critical role of boundary problems, we aim to maximize their exploitation through targeted re-sampling. 
Specifically, given an initial sampling budget of $n_t$ per query, we subsequently assign an additional sampling count $n_r$ to each boundary sample estimated by our RDE during the Re-Sampling phase, calculated as follows:
\begin{align}
n_r = \left[ \frac{(n_t - n_p) \times |\mathcal{D}_t|}{|\mathcal{D}_t^{\mathcal{B}}|} \right],
\end{align}
where $\mathcal{D}_t^{\mathcal{B}}$ represents the subset of samples classified as \emph{Boundary} in the Difficulty Estimation phase.
This reallocation of the sampling budget enables us to focus computational resources on such instances, which offer greater potential for optimization.
Subsequently, for each query $x_b\in\mathcal{D}_t^{\mathcal{B}}$, we utilize the policy model $\mathcal{M}_t$ to generate $n_r$ candidate responses, forming $\mathcal{R}_{t,x_b}^r = \{r_{1}, r_{2}, \ldots, {r}_{n_r} \mid {r}_i \sim \mathcal{M}_t(x_b) \}$.

\subsection{\textit{Phase 3}: Preference Optimization}

Based on the sampled responses from the aforementioned two phases, we construct a preference dataset to facilitate self-training via preference optimization.
At iteration $t$, for each query $x$, the policy model $\mathcal{M}_t$ has generated a response set $\mathcal{R}_{t,x} = {\mathcal{R}}_{t,x}^p \cup {\mathcal{R}}_{t,x}^r$. 
To construct the preference dataset $\mathcal{D}_t^{\text{pairs}}$, these responses are systematically categorized based on their correctness. For each query $x$, the response set $\mathcal{R}_{t,x}$ is partitioned into two subsets: the set of correct responses $\mathcal{R}_{t,x}^{\text{corr}} = \{r \in \mathcal{R}_{t,x} \mid V(x,r) = 1\}$, and the set of incorrect responses $\mathcal{R}_{t,x}^{\text{incorr}} = \{r \in \mathcal{R}_{t,x} \mid V(x,r) = 0\}$.

Subsequently, the samples in sets $\mathcal{R}_{t,x}^{\text{corr}}$ and $\mathcal{R}_{t,x}^{\text{incorr}}$ are independently ranked in descending order according to their reward scores $S(r)$. 
This produces the ordered sequences $\widetilde{\mathcal{R}}_t^{\text{corr}} = (r_{(1)}^{\text{corr}}, r_{(2)}^{\text{corr}}, \ldots, r_{(|\mathcal{R}_t^{\text{corr}}|)}^{\text{corr}})$ and $\widetilde{\mathcal{R}}_t^{\text{incorr}} = (r_{(1)}^{\text{incorr}}, r_{(2)}^{\text{incorr}}, \ldots, r_{(|\mathcal{R}_t^{\text{incorr}}|)}^{\text{incorr}})$. 
The number of pairs $k$ for $\mathcal{D}_t^{\text{pairs}}$ is defined as the minimum cardinality of these two sets: $k = \min (|\mathcal{R}_t^{\text{corr}}|, |\mathcal{R}_t^{\text{incorr}}|)$. 
Finally, the paired dataset is constructed as $\mathcal{D}_t^{\text{pairs}} = \{ (s_i^{\text{corr}}, s_i^{\text{incorr}}) \mid i=1, \ldots, k \}$, where each $s_i^{\text{corr}}$ and $s_i^{\text{incorr}}$ is a unique sample from the top $k$ elements of $\mathcal{R}_t^{\text{corr}}$ and $\mathcal{R}_t^{\text{incorr}}$, respectively.

By training on the given set of preference pairs, we derive the updated model $\mathcal{M}_{t+1}$, initialized from its predecessor $\mathcal{M}_t$. The optimization follows the DPO objective~\cite{rafailov2023direct}, as specified in Eq.~\ref{dpo_loss}.

\begin{table*}[t]
\centering
\renewcommand\arraystretch{1}
\setlength{\tabcolsep}{12pt}
\caption{
Main results across mathematical reasoning benchmarks. All \textsc{STaR}-based methods are trained iteratively for three self-training rounds. \textbf{Bold} values indicate the best performance, while \underline{underlined} ones denote the second-best results. For AMC23 and AIME24, we report Avg@32, and Pass@1 is used for others. 
}
\resizebox{1\linewidth}{!}{
\begin{tabular}{lcccccccc}
\toprule
\multicolumn{1}{c}{\textbf{Method}} & \textbf{GSM8K} & \textbf{\begin{tabular}[c]{@{}c@{}}MATH\\ 500\end{tabular}} & \textbf{\begin{tabular}[c]{@{}c@{}}Olympiad\\Bench \end{tabular}} & \textbf{\begin{tabular}[c]{@{}c@{}}Minerva\\ Math\end{tabular}} & \textbf{\begin{tabular}[c]{@{}c@{}}AMC23 \end{tabular}} & \textbf{\begin{tabular}[c]{@{}c@{}} College\\Math\end{tabular}} & \textbf{AIME24} & \textbf{Avg.} \\ 
\midrule
\textit{DeepSeek-Math-7B} & 30.3 & 18.6 & 5.3 & 5.9 & 7.3 & 17.2 & 0.0 & 12.1 \\
Vanilla SFT & 54.4 & 28.4 & 9.9 & 7.0 & 10.9 & \underline{28.4} & 0.6 & 19.9  \\ %
\hdashline
Stepwise Init. & 62.9 & 32.8 & \underline{10.7} & 8.1 & 11.3 & 25.4 & 0.4 & 21.7 \\ %
\quad+\textsc{STaR}-RFT & \underline{66.0} & 30.2 & 8.6 & \underline{11.8} & 11.7 & 26.8 & 0.5 & 22.2\\ %
\quad+\textsc{STaR}-DPO & 63.1 & \underline{33.8} & 9.9 & 10.7 & \underline{12.6} & 26.7 & \underline{0.7} & \underline{22.5}\\ %
\rowcolor{ourscolor}\quad+\textsc{HS-STaR}~(Ours) & \textbf{67.7} & \textbf{35.4} & \textbf{12.0} & \textbf{13.6} & \textbf{13.4} & \textbf{29.8} & \textbf{1.1} & \textbf{24.7} \\
\midrule
\textit{Phi-3.5-Mini-Instruct} & 83.5 & 46.2 & 13.2 & 16.2 & 16.2 & 36.1 & 0.8 & 30.3\\
Vanilla SFT & 81.7 & \underline{47.8} & 14.7 & 11.4 & 15.8 & 32.0 & 0.5 & 29.1 \\ %
\hdashline
Stepwise Init. & 85.4 & 45.2 & 13.5 & \underline{24.3} & 16.2 & 35.9 & \underline{1.2} & 31.7\\ %
\quad+\textsc{STaR}-RFT & 84.9 & 45.2 & \underline{15.6} & 23.9 & \underline{16.6} & 36.1 & 1.1 & 31.9 \\ %
\quad+\textsc{STaR}-DPO & \textbf{86.5} & 46.4 & 14.8 & \textbf{24.6} & 16.2 & \underline{36.2} & 0.8 & \underline{32.2} \\ %
\rowcolor{ourscolor}\quad+\textsc{HS-STaR}~(Ours)  & \underline{86.1} & \textbf{49.2} & \textbf{15.7} & \underline{24.3} & \textbf{17.4} & \textbf{36.5} & \textbf{1.9} & \textbf{33.0} \\
\midrule
\textit{Qwen2.5-3B} & 72.9 & 49.4 & 16.3 & 17.3 & 21.1 & 33.8 & 2.6 & 30.5\\
Vanilla SFT & 62.9 & \underline{58.6} & \textbf{23.6} & 13.2 & 25.5 & 31.0 & \textbf{3.6} & 31.2 \\ %
\hdashline
Stepwise Init. & 72.8 & 50.2 & 19.6 & 16.9 & 20.8 & 35.4 & 2.8 & 31.2\\ %
\quad+\textsc{STaR}-RFT & 80.1 & 51.4 & 19.6 & 19.1 & \underline{28.2} & \underline{39.7} & 1.6 & 34.2 \\ %
\quad+\textsc{STaR}-DPO & \underline{81.0} & 55.2 & 21.6 & \textbf{24.6} & 28.0 & 39.1 & 2.5 & \underline{36.0} \\ %
\rowcolor{ourscolor}\quad+\textsc{HS-STaR}~(Ours) & \textbf{82.6} & \textbf{60.0} & \underline{22.7} & \underline{24.3} & \textbf{28.4} & \textbf{40.7} & \underline{3.0} & \textbf{37.4} \\ %
\midrule
\textit{Qwen2.5-7B} & 81.8 & 54.2 & 25.6 & 25.4 & 26.4 & 39.3 & 3.7 & 36.6 \\
Vanilla SFT & 84.2 & 66.8 & 25.9 & 17.3 & 37.0 & 36.8 & 6.9 & 39.3\\ %
\hdashline
Stepwise Init. & 86.4 & 65.0 & 27.9 & 25.0 & 35.0 & 41.7 & 5.2 & 40.9 \\ %
\quad+\textsc{STaR}-RFT & 86.7 & 66.8 & 27.2 & \underline{31.4} & \underline{45.2} & 38.7 & 4.8 & 43.0 \\ %
\quad+\textsc{STaR}-DPO & \underline{88.6} & \underline{69.8} & \underline{33.3} & 29.8 & 44.3 & \underline{45.7} & \underline{8.3} & \underline{45.7} \\ %
\rowcolor{ourscolor}\quad+\textsc{HS-STaR}~(Ours)  & \textbf{90.3} & \textbf{72.8} & \textbf{35.9} & \textbf{31.6} & \textbf{46.5} & \textbf{46.4} & \textbf{8.9} & \textbf{47.5} \\ %
\bottomrule
\end{tabular}
}
\label{tab:generalize_results}
\end{table*}

\section{Experiments}
\label{sec:experiments}
\subsection{Setup}
\label{sec: setup}
\noindent\textbf{Dataset.}
Following~\citet{zhangonline}, we use NuminaMath-1.5~\cite{numina_math_15_datasets} for iterative self-taught reasoning.  
The original dataset contains approximately 900K math problems, and we apply a filtering pipeline to ensure the quality of questions and the verifiability of answers.  
In each iteration, we randomly sample 7{,}500 problems without replacement, ensuring no overlap across iterations.  
Additional details are provided in Appendix~\ref{sec:appendix-dataset}.

\noindent\textbf{Implementation Details.}
To facilitate the generation of stepwise solutions for reward labeling, we first perform a \textbf{warm-up training} using synthetic solutions. Specifically, we leverage the MATH dataset~\cite{hendrycks2021measuring} and prompt \textit{gpt-4o-2024-08-06} to systematically rewrite each solution in a step-by-step format, then organize these steps separated by "\verb|\n\n|". The resulting model, denoted as $\mathbf{M}_0$, serves as the initialization for iterative self-training. 
In our experiments, each iteration operates under a fixed sampling budget, corresponding to an average of 8 samples per problem. The pre-sampling count $n_p$ is set to 3, and thresholds $\tau_h$ and $\tau_l$ for difficulty estimation are set to 0.15 and 0.65, respectively. We utilize Skywork-PRM-7B~\cite{skyworkopeno12024} as our PRM and perform three iterations in total. See more details in Appendix~\ref{sec:appendix-implementation}.

\begin{table*}[t]
\centering
\renewcommand\arraystretch{1}
\setlength{\tabcolsep}{12pt}
\caption{Comparison with other ``Zero Training'' models. All models are fine-tuned based on the Qwen2.5-Math-7B. We evaluate SimpleRL, PURE-VR, and DPO-VP using their publicly released checkpoints, while \textsc{STaR}-RFT and \textsc{STaR}-DPO are reproduced under the same experimental settings as ours.}

\resizebox{1\linewidth}{!}{
\begin{tabular}{lcccccccc}
\toprule
\multicolumn{1}{c}{\textbf{Method}} & \textbf{\begin{tabular}[c]{@{}c@{}}Training\\Strategy \end{tabular}} & \textbf{\begin{tabular}[c]{@{}c@{}}MATH\\ 500\end{tabular}} & \textbf{\begin{tabular}[c]{@{}c@{}}Olympiad\\Bench \end{tabular}} & \textbf{\begin{tabular}[c]{@{}c@{}}Minerva\\ Math\end{tabular}} & \textbf{\begin{tabular}[c]{@{}c@{}}AMC23 \end{tabular}} & \textbf{\begin{tabular}[c]{@{}c@{}} College\\Math\end{tabular}} & \textbf{AIME24} & \textbf{Avg.} \\ 
\midrule

Qwen2.5-Math-7B & - & 72.0 & 34.8 & 27.6 & 56.1 & 43.0 & 17.2 & 41.8 \\
Qwen2.5-Math-7B-Instruct & - & \textbf{82.8} & 40.3 & 35.7 & 59.5 & 46.9 & 11.4 & 46.1 \\
\hdashline
SimpleRL & online RL &  78.2 & \textbf{42.5} & 34.2 & 62.3 & \textbf{49.1} & \textbf{23.9} & \textbf{48.4} \\
PURE-VR & online RL & \underline{79.0} & 40.6 & \textbf{36.4} & \underline{63.1} & 47.3 & 15.6 & 47.0 \\
DPO-VP & STaR & 74.4 & 36.4 & 31.2 & 57.5 & 45.1 & 18.9 & 43.9 \\
\textsc{STaR}-RFT & STaR & 73.8 & 37.9 & 36.0 & 62.3 & 47.0 & 18.3 & 45.9 \\ %
\textsc{STaR}-DPO & STaR & 77.6 & 41.3 & 34.9 & 60.5 & \underline{48.0} & 19.5 & 47.0 \\ %
\rowcolor{ourscolor} \textsc{HS-STaR}~(Ours) & STaR & 77.8 & \underline{41.8} & \textbf{36.4} & \textbf{64.3} & \underline{48.0} & \underline{20.8} & \underline{48.2} \\  
\bottomrule
\end{tabular}
}
\label{tab:perf-qwen2.5-math-7b-wo-gsm}
\end{table*}

\noindent\textbf{Baselines.}
To ensure a comprehensive evaluation, we apply \textsc{HS-STaR} across a diverse set of open-source models, including DeepSeek-Math-7B~\cite{shao2024deepseekmathpushinglimitsmathematical}, Phi-3.5-Mini-Instruct~\cite{abdin2024phi3technicalreporthighly}, Qwen2.5-3B, and Qwen2.5-7B~\cite{qwen2025qwen25technicalreport}. We compare with the following baselines: (1) \textbf{Vanilla SFT}, using reference solutions from NuminaMath for training without any self-generated data; (2) \textbf{Stepwise Initialization ($\mathbf{M}_0$)}, the base model trained on synthetic step-by-step solutions without any self-training; 
(3) \textbf{\textsc{STaR}-RFT}, using SFT as the training objective in STaRs; and (4) \textbf{\textsc{STaR}-DPO}, using DPO as the training objective in STaRs.

\noindent\textbf{Evaluation.}
We evaluate our framework on seven mathematical reasoning benchmarks, including GSM8K~\cite{cobbe2021gsm8k}, MATH500~\cite{yang2024qwen25mathtechnicalreportmathematical}, OlympiadBench~\cite{he-etal-2024-olympiadbench}, Minerva-Math~\cite{lew-minerva-math}, College-Math~\cite{tang-math-scale}, as well as competition-level benchmarks such as AMC23~\cite{amc23} and AIME24~\cite{aime24}.  
We report \textbf{Pass@1} accuracy for all benchmarks, with the exception of AMC23 and AIME24. For these two, we follow standard protocol and report \textbf{Avg@32}, which is calculated from 32 generated samples per problem, using temperature as 0.6.

\subsection{Main Results}
Table~\ref{tab:generalize_results} presents a comparative study of training methods across multiple mathematical reasoning benchmarks and backbone LLMs. We can draw the following conclusions:

\noindent\textbf{\textsc{HS-STaR} achieves superior performance.}
Across all model backbones and benchmarks, \textsc{HS-STaR} consistently outperforms baseline methods. For example, it improves the overall accuracy by 2.2\% on DeepSeek-Math-7B, 1.4\% on Qwen2.5-3B, and 1.8\% on Qwen2.5-7B compared to their respective best-performing baselines. These results demonstrate the significance of identifying and exploiting high-utility problems. Furthermore, on challenging datasets such as AIME24 and AMC23, \textsc{HS-STaR} also outperforms the most competitive counterparts, demonstrating the robustness of our boundary-focused sampling strategy.

\noindent\textbf{DPO consistently outperforms RFT.}  
Across most settings, \textsc{STaR}-DPO achieves higher accuracy than \textsc{STaR}-RFT. For instance, on Qwen2.5-7B and Qwen2.5-3B, \textsc{STaR}-DPO yields relative gains of 2.7\% and 1.8\%, respectively.  
We assume that this stems from DPO’s ability to leverage both correct and incorrect responses, whereas RFT relies solely on correct trajectories and may underutilize informative failure cases.

\noindent\textbf{Iterative self-training brings improvements.}  
We observe that all STaR-based methods consistently outperform their initializations and vanilla SFT baselines, validating the effectiveness of the training paradigm.  
Among the backbones, the relatively modest improvement observed on Phi-3.5-Mini-Instruct is likely due to the extensive post-training it has already undergone.  
Moreover, we find that stepwise initialization not only enables format-consistent reasoning but also outperforms vanilla SFT, demonstrating its effectiveness as a lightweight and generalizable warm-up strategy.

\begin{figure}[t]
\centering
\includegraphics[width=\linewidth]{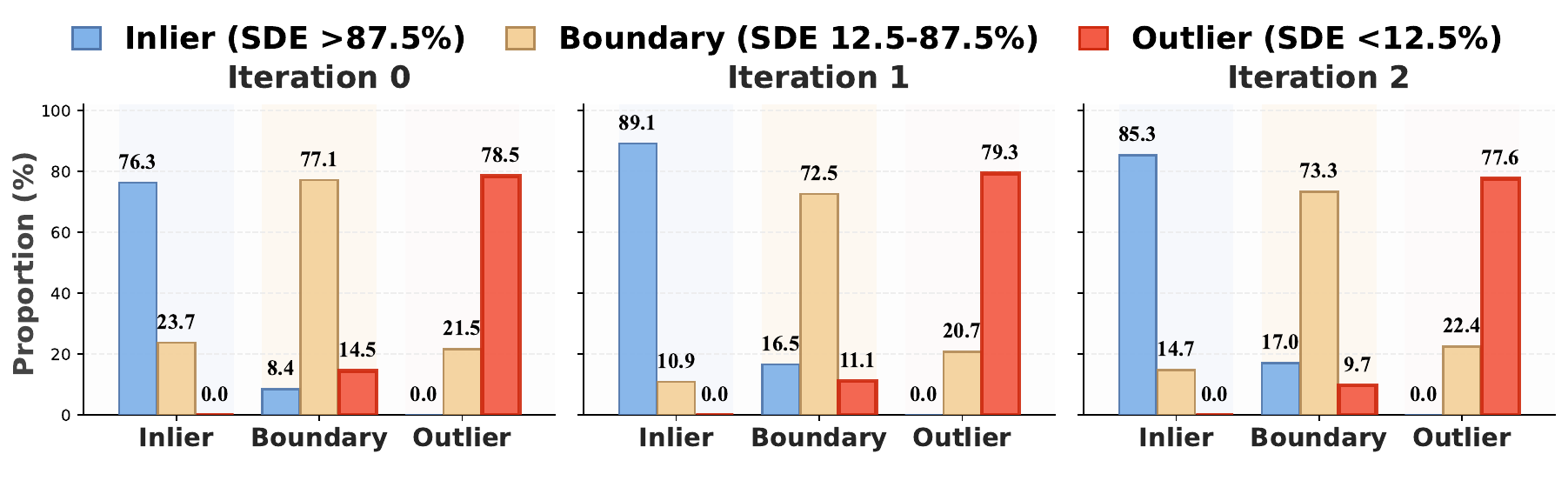}
\caption{Estimation performance of our RDE on Qwen2.5-7B model. Sample categories identified by ours are presented along the horizontal axis, and for each category, the vertical dimension indicates the proportion of samples belonging to that category as estimated by SDE.}
\label{fig:estimation}
\end{figure}

\begin{figure}[t]
\centering
\includegraphics[width=\linewidth]{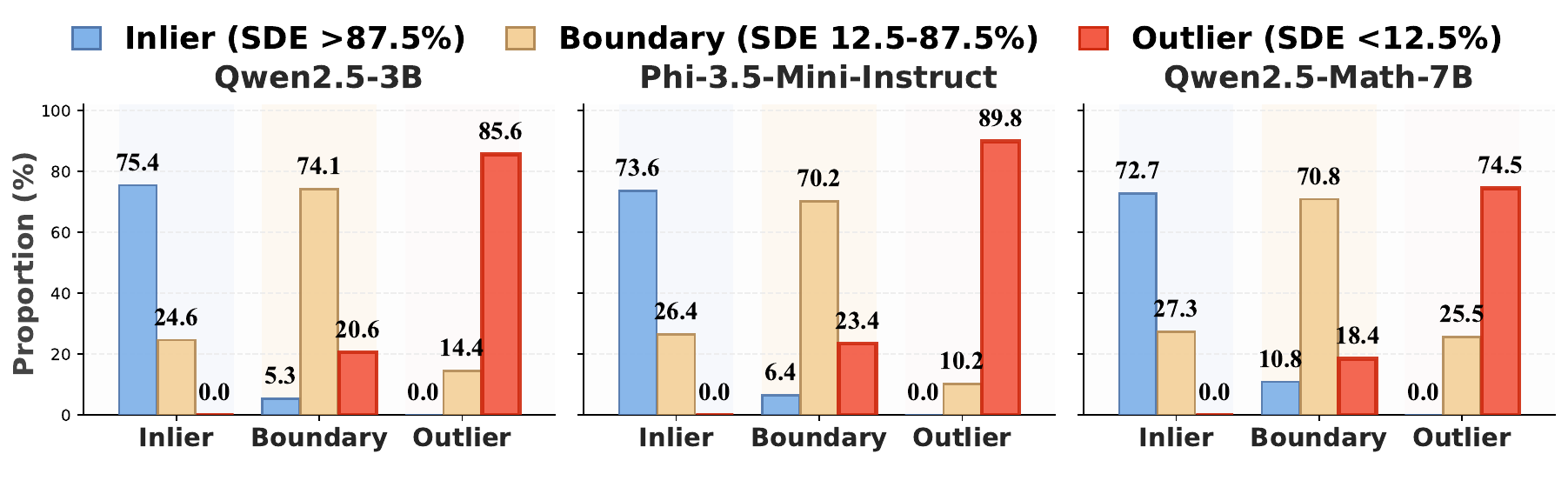}
\caption{Estimation performance of our RDE on Qwen2.5-3B, Phi3.5-Mini-Instruct, and Qwen2.5-Math-3B. Notably, Qwen2.5-Math-7B is evaluated under the ``Zero Training'' setting.}
\label{fig:estimation_three_model}
\end{figure}

\subsection{``Zero Training'' of \textsc{HS-STaR}}

\noindent\textbf{Settings.} The recent emergence of DeepSeek-R1~\cite{deepseekai2025deepseekr1incentivizingreasoningcapability} has sparked a trend of R1-Zero-like training \cite{chu2025gpg,yu2025dapoopensourcellmreinforcement}, where reinforcement learning is applied directly to pre-trained models. Following this, we explore a similar ``Zero Training'' setup to further evaluate our approach. Specifically, we conduct \textsc{HS-STaR} on Qwen2.5-Math-7B~\cite{yang2024qwen25mathtechnicalreportmathematical} by skipping the initial warm-up SFT. 
We compare against various advanced LLM reasoning training methods over the same backbone, including 
Qwen2.5-Math-7B-Instruct~\cite{yang2024qwen25mathtechnicalreportmathematical}, SimpleRL~\cite{zeng2025simplerlzooinvestigatingtamingzero}, PURE-VR~\cite{cheng2025stopsummationminformcredit}, DPO-VP~\cite{tu2025enhancingllmreasoningiterative}, \textsc{STaR}-RFT (named Online RFT in \citet{zeng2025bstar}), and \textsc{STaR}-DPO (referred to as Online DPO in~\citet{zhangonline}).
Detailed descriptions of these methods are provided in Appendix~\ref{sec:appendix-baselines}.

\noindent\textbf{Results.} As illustrated in Table~\ref{tab:perf-qwen2.5-math-7b-wo-gsm}, we observe that \textsc{HS-STaR} also demonstrates strong performance in zero training setting, achieving a 6.4\% improvement over the backbone model. Among self-training approaches, \textsc{HS-STaR} achieves the highest accuracy, surpassing the second-best method by 1.2\%. Moreover, we find that \textsc{HS-STaR} can even achieve performance comparable to SimpleRL, which leverages GRPO~\cite{shao2024deepseekmath} for reinforcement learning. This suggests that the proposed framework can match the performance of online RL through a more flexible framework, while avoiding the complexity of hyperparameter tuning and the computational costs~\cite{abdin2024phi3technicalreporthighly, tu2025enhancingllmreasoningiterative, liu2025understandingr1zeroliketrainingcritical, fu2025rlaereinforcementlearningassistedensemble, wang2025emergent}.

\subsection{Analysis of Difficulty Estimation}
\label{analysis_on_mde}

\textbf{Impact on Estimation Strategy of \textsc{HS-STaR}.} 
Our \textsc{HS-STaR} can be seamlessly integrated with several alternative approaches to difficulty estimation. Specifically, we have developed three variants: \textsc{HS-STaR}-Acc, which solely utilizes accuracy-based estimation, \textsc{HS-STaR}-Reward, which solely utilizes reward-based estimation, and \textsc{HS-STaR}-SDE, which employs SDE (defined in Sec.~\ref{sec: preliminary}) as an oracle measure of problem difficulty.
Further details of these variants are provided in Appendix~\ref{sec:appendix-baselines}.
As summarized in Table~\ref{tab:ablation-on-estimation_variants}, \textsc{HS-STaR}-SDE, which uses oracle difficulty and allocates more samples to boundary-level problems, leads to the best overall performance across all iterations, confirming the high utility of such problems. 
Among all variants that employ the same resource constraints, \textsc{HS-STaR} performs best, with accuracy only 0.4\% lower than the \textsc{HS-STaR}-SDE oracle. 
In contrast, both ablation variants result in noticeable performance drops, yet still outperform the naive \textsc{STaR}-DPO without difficulty estimation and budget reallocation. These results suggest that RDE offers an effective solution for difficulty estimation by combining two complementary signals, without requiring extensive sampling.

\noindent\textbf{Estimation Accuracy.} To evaluate the performance of our difficulty estimation method, we employ the labels derived by the SDE as the ground truth. As illustrated in Fig.~\ref{fig:estimation}, our method achieve an estimation accuracy on the three types of samples, exceeding 70\% across three iterations conducted on the Qwen2.5-7B model. 
Furthermore, as shown in Fig.~\ref{fig:estimation_three_model}, our method show considerable effectiveness across various models.
Notably, it maintains high accuracy even when evaluated on the Qwen2.5-Math-7B model trained under ``Zero Training'' settings.

\begin{table}[t]
\centering
\renewcommand\arraystretch{1}
\setlength{\tabcolsep}{8pt}
\caption{Ablation study on difficulty estimation. We report the average performance across seven benchmarks.}
\resizebox{0.85\linewidth}{!}{
\begin{tabular}{lccc}
\toprule
\textbf{Method} & \textbf{Iter. 1} & \textbf{Iter. 2} & \textbf{Iter. 3} \\
\midrule
\textsc{HS-STaR}-SDE (Oracle) & \textbf{45.7} & \textbf{47.0} & \textbf{47.9} \\
\hdashline
\textsc{STaR}-DPO & 44.2 & 45.1 & 45.7 \\
\textsc{HS-STaR}-Acc & 44.7 & 46.2 & 46.8\\
\textsc{HS-STaR}-Reward & 45.4 & 46.3 & 46.7\\
\rowcolor{ourscolor} \textsc{HS-STaR} & \underline{45.6} & \underline{46.6} & \underline{47.5} \\
\bottomrule
\end{tabular}
}
\label{tab:ablation-on-estimation_variants}
\end{table}

\begin{figure}[t]
\centering
\begin{subfigure}{0.48\linewidth}
    \includegraphics[width=\linewidth]{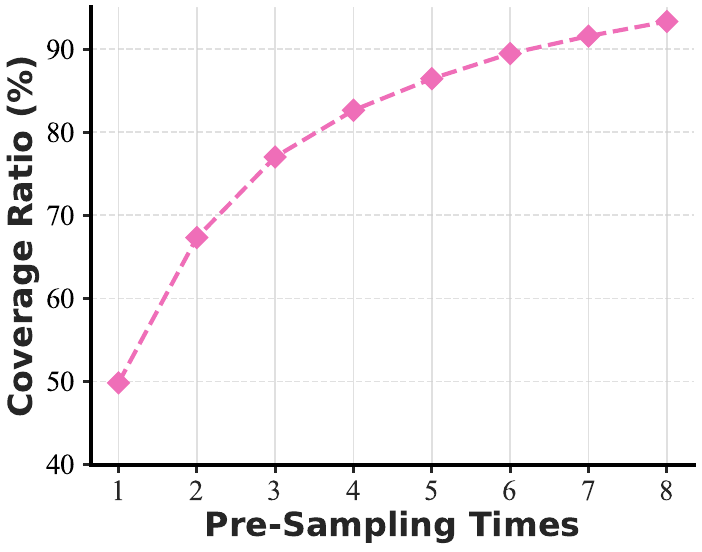}
    \subcaption{Boundary Samples Coverage in Re-Sampling Stage.}
    \label{fig:presampling-on-coverage}
\end{subfigure}
\hfill
\begin{subfigure}{0.48\linewidth}
    \includegraphics[width=\linewidth]{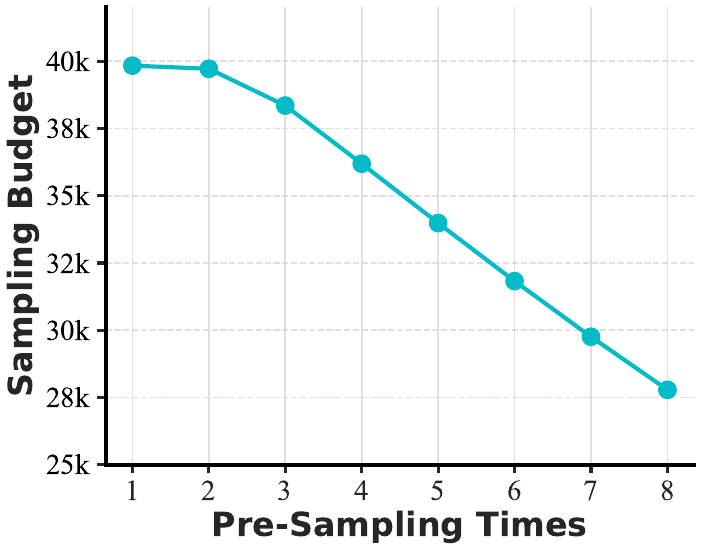}
    \subcaption{Total Sampling Budget on Boundary Samples.}
    \label{fig:presampling-on-budget}
\end{subfigure}
\caption{Analysis of the effects of Pre-Sampling times on Qwen2.5-7B. Subfig.~\ref{fig:presampling-on-coverage} shows the trend of coverage of SDE estimated boundary samples as Pre-Sampling times vary. 
Subfig.~\ref{fig:presampling-on-budget} illustrates how the total sampling budget for SDE estimated boundary samples evolves as Pre-Sampling times vary.
}
\label{fig:effect-presampling}
\end{figure}

\begin{figure}[t]
\centering
\includegraphics[width=\linewidth]{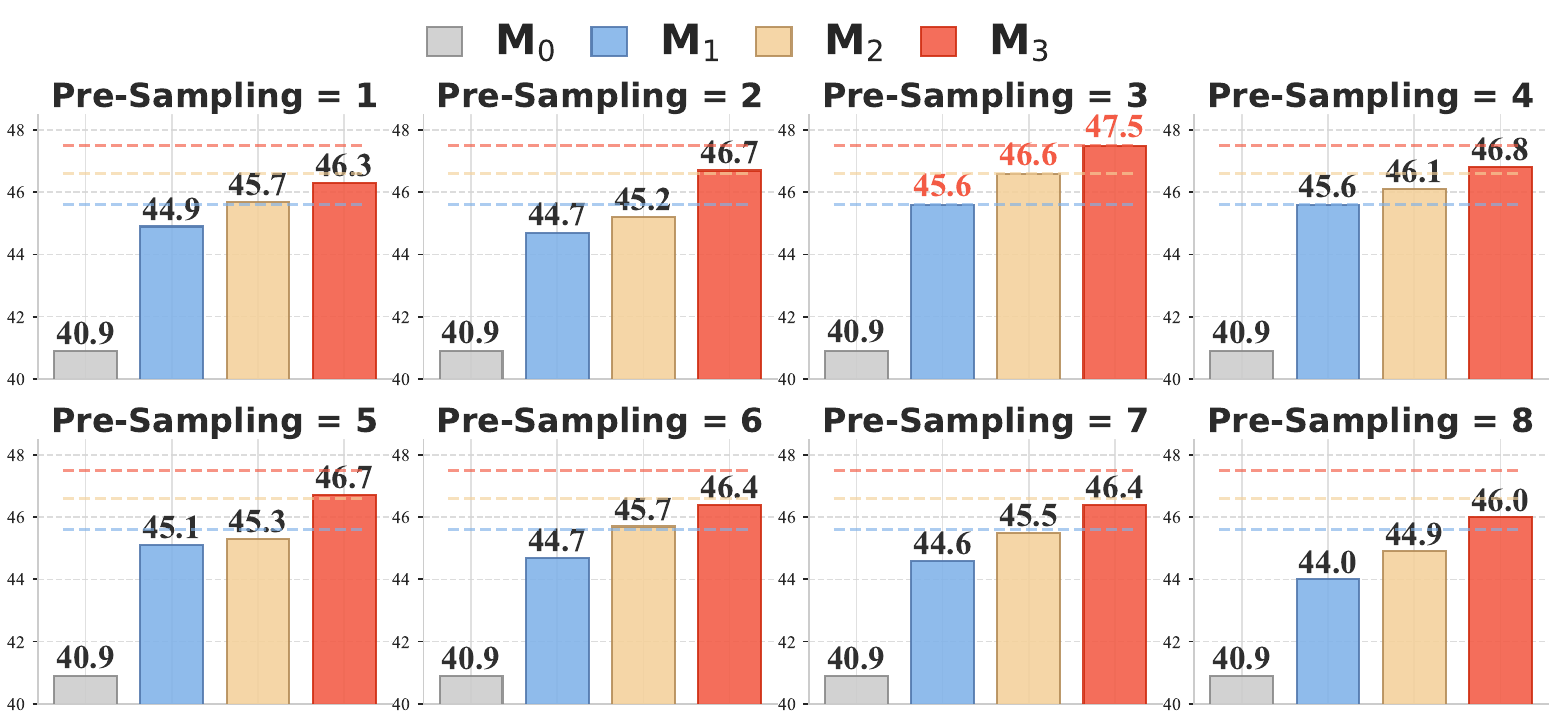}
\caption{
Performance under different Pre-Sampling Times on Qwen2.5-7B. For comparative analysis, we use the performance at a Pre-Sampling Time of 3 as the baseline, indicated by the dashed line.}
\label{fig:presampling-on-perf}
\end{figure}

\subsection{Impact of Pre-Sampling Times}

Since the number of pre-sampling directly influences both estimation accuracy and budget allocation, we examine its impact in detail.
As shown in Fig.~\ref{fig:presampling-on-coverage}, increasing the number of pre-sampling times improves the accuracy of difficulty estimation, leading to better coverage of Boundary samples. 
However, this gain in estimation accuracy introduces a trade-off. As depicted in Fig.~\ref{fig:presampling-on-budget}, under a fixed total sampling budget, allocating more resources to pre-sampling reduces the budget available for exploiting high-utility Boundary samples, weakening overall reallocation effectiveness. 
This tradeoff is further confirmed in Fig.~\ref{fig:presampling-on-perf}, which shows that performance peaks when pre-sampling is performed three times, balancing estimation accuracy and budget efficiency.  
Beyond this point, further increasing pre-sampling leads to performance degradation due to insufficient sampling of critical instances.

\begin{table}[t]
\centering
\renewcommand\arraystretch{1}
\setlength{\tabcolsep}{8pt}
\caption{Impact of threshold ranges on precision, recall, and average performance. Our chosen setting strikes a balance, leading to the highest average performance.}
\resizebox{0.85\linewidth}{!}{
\begin{tabular}{lccc}
\toprule
\textbf{Configuration} & \textbf{Precision} & \textbf{Recall} & \textbf{Avg.} \\
\midrule
Narrower Range         & 81.6                   & 67.8                & 47.0                      \\
Broader Range          & 52.9                   & 92.5                & 46.6                      \\
\rowcolor{ourscolor} Standard          & 77.1                   & 73.8                & 47.5                      \\
\bottomrule
\end{tabular}
}
\label{tab:threshold_impact}
\end{table}

\subsection{Sensitive Analysis of Thresholds}
In our reward-guided difficulty estimation approach, the hyperparameters $\tau_l$ and $\tau_h$ are pivotal for defining and selecting boundary cases. These thresholds are empirically determined through pilot experiments, as detailed in Sec.~\ref{sec:pilot}. Importantly, this calibration constitutes a one-time overhead for a given process reward model. The resulting thresholds generalize robustly across all policy models in our main experiments, thereby obviating the need for model-specific fine-tuning.

Our selection process was primarily guided by the trade-off between precision and recall in identifying these boundary samples. Using the boundary cases identified by the SDE method as the ground truth, we define Precision as the fraction of samples classified as boundary cases by our method that are also identified as such by the SDE method, and Recall as the fraction of all boundary cases identified by the SDE method that our method successfully detects. These two metrics dictate the balance between the sampling budget allocated to boundary cases and the diversity of these cases captured during the Re-Sampling phase.

To further validate our choice and elucidate the impact of these thresholds, we conducted an ablation study that considered a narrower range by setting $\tau_l = 0.4$ and $\tau_h = 0.6$, as well as a broader range with $\tau_l = 0.2$ and $\tau_h = 0.8$.
As the results in Table~\ref{tab:threshold_impact} indicate, a clear trade-off emerges. A narrower range yields higher precision but at the cost of significantly lower recall. Conversely, a broader range substantially increases recall, but this is achieved at the expense of a sharp decline in precision. These experimental results confirm that our chosen thresholds achieve an effective balance between precision and recall, and that this equilibrium is conducive to better overall performance.

\begin{table}[t]
\centering
\renewcommand\arraystretch{1}
\setlength{\tabcolsep}{10pt}
\caption{Ablation study of Re-Sampling strategies on Qwen2.5-7B. For these variants, the estimation of Inlier, Boundary, and Outlier samples is performed using RDE.}
\resizebox{0.9\linewidth}{!}{
\begin{tabular}{lccc}
\toprule
\textbf{Re-Sampling  Strategy} & \textbf{Iter.~1} & \textbf{Iter.~2} & \textbf{Iter.~3} \\
\midrule
w/o Re-Sampling & 44.2 & 45.1 & 45.7\\
Inlier & 42.0 & 43.0 & 44.2 \\
Outlier & 41.7 & 41.9 & 42.2 \\
Inlier+Outlier & 41.9 & 41.5 & 42.8 \\
Inlier+Boundary & \underline{45.3} & \underline{46.4} & \underline{47.2} \\
Boundary+Outlier & 44.2 & 44.7 & 46.0 \\
\rowcolor{ourscolor}Boundary~(Ours) & \textbf{45.6} & \textbf{46.6} & \textbf{47.5} \\
\bottomrule
\end{tabular}
}
\label{tab:ablation-on-samples-selection-avg}
\end{table}

\subsection{Comparison of Re-Sampling Strategies}

In \textsc{HS-STaR}, the re-sampling budget is allocated based on difficulty levels estimated by our RDE strategy, with a focus on boundary-level problems.  
To better assess the utility of different difficulty levels, we compare re-sampling strategies constructed from all possible combinations of \textit{Inlier}, \textit{Outlier}, and \textit{Boundary} samples, excluding the boundary-only configuration used in our main approach.
The results are presented in Table~\ref{tab:ablation-on-samples-selection-avg}, where re-sampling exclusively on \textit{Boundary} samples consistently yields the best performance across all iterations, confirming the effectiveness of prioritizing such problems to maximize training utility.
Strategies involving \textit{Inlier+Boundary} also perform competitively, likely due to the predominance of boundary samples in the combined set.  
In contrast, strategies based on \textit{Inlier}, \textit{Outlier}, or their combinations result in significantly lower performance.  
These findings highlight the importance of focusing on boundary-level queries during the re-sampling stage for effective self-improvement.

\section{Related Work}

\subsection{Self-Taught Reasoners}
Recent studies have shown that LLMs can progressively improve themselves by training on self-generated responses using SFT or DPO~\cite{zelikman2022star,gulcehre2023_rest,huang-etal-2023-lmsi,yuan2024_self_rewarding,li202512surveyreasoning, wang2025positionbiasmitigatesposition}. In mathematical reasoning tasks, response selection is typically guided by answer correctness, enabling LLMs to act as self-taught reasoners without relying on human-annotated reasoning trajectories~\cite{yuan2023_rejection_sampling_finetune,singh2024beyond,hosseini2024vstar,pang2024iterative,wu2025_progress_or_regress,zhangonline}.

Previous research has primarily explored two further directions. One line of work incorporates auxiliary reward model signals beyond answer correctness~\cite{yang2024qwen25mathtechnicalreportmathematical,zeng2025bstar,tu2025enhancingllmreasoningiterative}. Another focuses on enhancing the quality or accuracy of sampled responses, including designing MCTS strategy~\cite{zhang2024restmcts,tian2024_alphallm,chen2024alphamath,wang2024_alphallm_cpl} and integrating teacher guidance~\cite{ding-etal-2025-mitigating-gsi}. 
However, these methods mainly aim to improve response quality, without accounting for the varying utility of problems. In contrast, our study reveals that boundary-level problems play a pivotal role in self-taught reasoning and introduces a hierarchical sampling strategy to efficiently exploit their utility. 

\subsection{Difficulty-Aware LLM Training}

Difficulty-aware strategies have proven effective for improving the training of LLMs. For instance, in instruction tuning, prior work commonly adopts instruction-following difficulty~\cite{li-etal-2024-quantity,li-etal-2024-selective} or uncertainty-based techniques~\cite{liu2024selectit,zhang2025d3diversitydifficultydependabilityaware} to select high-utility data.  
In mathematical reasoning, problem difficulty is typically estimated by pass rate. On this basis, DART-MATH~\cite{tong_dart_math} allocates more sampling budget to synthesize hard examples, while some recent studies advocate avoiding overly difficult questions~\cite{tian2025deepdistillenhancingllmreasoning,bae2025onlinedifficultyfilteringreasoning,yu2025dapoopensourcellmreinforcement}. 
Within STaR, a few studies have explored difficulty-aware sampling by allocating more resources to challenging problems~\cite{ding-etal-2025-mitigating-gsi,xue2025dastdifficultyawareselftraininglarge}. 
However, our analysis demonstrates that such difficult questions contribute significantly less compared to those near the model's capability boundary. Therefore, we propose \textsc{HS-STaR} that efficiently identifies and prioritizes boundary-level problems during self-training.

\section{Conclusion}

In this paper, we empirically demonstrate that the utility of self-training data is largely determined by the difficulty level of problems, with problems near the model's capability boundary being substantially more valuable than overly simple or excessively hard ones. Motivated by these findings, we propose \textsc{HS-STaR}, a hierarchical sampling framework that improves self-taught reasoning by explicitly estimating and exploiting problem utility. 
Concretely, \textsc{HS-STaR} first performs lightweight reward-guided difficulty estimation, then reallocates the sampling budget to prioritize high-utility boundary-level problems for preference optimization, thereby maximizing training effectiveness under a fixed sampling resource constraint. 
Experimental results confirm that our method significantly outperforms various baselines.
We believe this work provides valuable insights for difficulty-aware optimization in LLM post-training.

\section*{Limitations}

Despite the proposed framework \textsc{HS-STaR} effectively enhances the self-training for mathematical reasoning, it has two primary limitations.

\begin{itemize}[leftmargin=*,itemsep=0.5pt, topsep=1.5pt]
    \item \textsc{HS-STaR} relies on difficulty estimation techniques such as reward-guided estimation to identify high-utility problems. Therefore, our framework is inherently tied to mathematical tasks, where problem difficulty is relatively well-defined. This limits the generalizability of \textsc{HS-STaR} to other domains where difficulty estimation is more ambiguous.

    \item Recent advances in rule-based RL have shown promising improvements in LLM reasoning. Although \textsc{HS-STaR} is developed for offline reinforced self-training, we believe that dynamically identifying high-utility problems during rollout could further improve the effectiveness of online RL, leaving this to our future work.
\end{itemize}

\noindent We believe that addressing these limitations could broaden the applicability of \textsc{HS-STaR}.

\bibliography{custom}

\appendix

\clearpage

\section{Experimental Settings}
This section offers comprehensive descriptions of the datasets, baselines, and implementations.

\subsection{Dataset Details}
\label{sec:appendix-dataset}
NuminaMath-1.5~\cite{numina_math_15_datasets} is the second iteration of the widely used NuminaMath~\cite{numina_math_datasets} dataset. This dataset provides a substantial collection of high-quality data suitable for post-training applications, comprising approximately 900,000 competition-level mathematics problems. Each problem is accompanied by a detailed solution presented in a Chain of Thought (CoT) format, which delineates the step-by-step reasoning process. The dataset encompasses a broad spectrum of mathematical content, drawing from diverse sources such as Chinese high school mathematics exercises and problems featured in prominent US and international mathematics olympiad competitions. The data collection primarily involved extracting content from online examination paper PDFs and mathematics discussion forums, thereby ensuring both the diversity and rigor of the included mathematical material.

\subsection{Implementation Details}
\label{sec:appendix-implementation}
We conducted both data sampling and model evaluation using the vLLM framework~\cite{kwon2023efficient}. 
During sampling, we set the temperature to 0.7. 
All models underwent full-parameter fine-tuning.
Specifically, we used a learning rate of $5 \times 10^{-7}$ for Qwen2.5-7B, while we trained Qwen2.5-Math-7B, Qwen2.5-3B, DeepSeek-Math-7B, and Phi-3.5-Instruct with a learning rate of $1 \times 10^{-6}$. 
Common hyperparameters included a maximum sequence length of 2048, a coefficient $\beta$ of 0.1, and a batch size of 256. For models incorporating a warmup phase, $\tau_l$ and $\tau_h$ were set to 0.15 and 0.65, respectively. In the "Zero-Training" scenario, $\tau_l$ and $\tau_h$ were assigned values of 0.15 and 0.4, respectively.

\subsection{Baseline Details}
\label{sec:appendix-baselines}

\subsubsection{Baselines on Qwen2.5-Math-7B}

\noindent\textbf{Qwen2.5-Math-7B-Instruct.}~\cite{qwen2025qwen25technicalreport} An instruct model in the Qwen2.5 series with strong mathematical reasoning capabilities.

\noindent\textbf{SimpleRL.}~\cite{zeng2025simplerlzooinvestigatingtamingzero} A reinforcement learning framework that enables zero-RL training from a base model, utilizing simple rule-based rewards to improve reasoning accuracy.

\noindent\textbf{PURE-VR.}~\cite{cheng2025stopsummationminformcredit} PURE is a reinforcement learning approach for LLM fine-tuning that replaces the standard sum-form credit assignment with a novel min-form, where the value function is defined as the minimum of future rewards.

\noindent\textbf{DPO-VP.}~\cite{tu2025enhancingllmreasoningiterative}
DPO-VP enhances LLM reasoning via iterative preference learning with DPO. It iteratively refines the generator and reward model using simple verifiable rewards, achieving efficient performance comparable to RL.

\noindent\textbf{\textsc{STaR}-RFT.} 
\textsc{STaR}-RFT is an iterative self-training method. At each iteration, it filters and selects correct answers based on their correctness to use for further training, thereby effectively achieving self-improvement.

\noindent\textbf{\textsc{STaR}-DPO.}
\textsc{STaR}-DPO is an iterative self-training method based on DPO. In each iteration, it partitions the generated responses based on their correctness and sorts the samples by reward to construct a preference dataset for DPO optimization.

\begin{algorithm}[t]
\small
\caption{\textsc{HS-STaR}}
\label{alg:main_framework}
\SetAlgoLined
\DontPrintSemicolon
\KwIn{Iterations $T$, initial policy model $\mathcal{M}_0$, dataset $\{\mathcal{D}_t\}_{t=1}^T$, sampling budgets $n_p$, $n_t$, functions $V(x,r)$, $S(x,r)$, thresholds $\tau_h$, $\tau_l$.}
\KwOut{Optimized model $\mathcal{M}_T$.}
\For{$t = 0$ \KwTo $T-1$}{
    Initialize $\mathcal{D}_t^{\text{pairs}} \leftarrow \emptyset$, $\mathcal{D}_t^{\mathrm{B}} \leftarrow \emptyset$\;
    \ForEach{$x \in \mathcal{D}_t$}{
        $\mathcal{R}_{t,x}^p \leftarrow \{r_{i} \sim \mathcal{M}_t(x)\}_{i=1}^{n_p}$\;
        Calculate $\phi_{\mathrm{a}}(\mathcal{R}_{t,x}^p) = \frac{1}{n_p} \sum_i V(x,r_i)$ and $\phi_{\mathrm{r}}(\mathcal{R}_{t,x}^p) = \frac{1}{n_p}\sum_i S(x,r_i)$\;
        \If{$x$ is classified as Boundary}{
            Add $x$ to $\mathcal{D}^{\mathrm{B}}$\;
        }
    }    
    \If{$|\mathcal{D}^{\mathrm{B}}| > 0$}{
        $n_r \leftarrow \left[ \frac{(n_t - n_p) \times |\mathcal{D}|}{|\mathcal{D}^{\mathrm{B}}|} \right]$\;
        \ForEach{$x \in \mathcal{D}^{\mathrm{B}}$}{
            Generate responses: $\mathcal{R}_{t,x}^r \leftarrow \{r_{i} \sim M_t(x)\}_{i=1}^{n_r}$\;
        }
    }

    \ForEach{$x \in \mathcal{D}^{\mathrm{B}}$}{
        $\mathcal{R}_{t,x} \leftarrow \mathcal{R}_{t,x}^p \cup \mathcal{R}_{t,x}^r$\;
        Partition $\mathcal{R}_{t,x}$ into $\mathcal{R}_{t,x}^{\text{corr}}$ and $\mathcal{R}_{t,x}^{\text{incorr}}$\;
        Sort $\mathcal{R}_{t,x}^{\text{corr}}$ and $\mathcal{R}_{t,x}^{\text{incorr}}$ in descending order\;
        $k \leftarrow \min(|\mathcal{R}_{t,x}^{\text{corr}}|, |\mathcal{R}_{t,x}^{\text{incorr}}|)$\;
        Sample $k$ pairs from top-$k$ responses: $\mathcal{D}_t^{\text{pairs}} \leftarrow \mathcal{D}_t^{\text{pairs}} \cup \{(r_{(i)}^{\text{corr}}, r_{(i)}^{\text{incorr}})\}_{i=1}^k$\;
    }
    Update $\mathcal{M}_{t+1}$ using DPO loss on $\mathcal{D}_t^{\text{pairs}}$\;
}
\Return{$\mathcal{M}_T$}
\end{algorithm}

\begin{table*}[ht]
\centering
\caption{Prompt Templates for Stepwise Solutions Construction.}
\resizebox{\linewidth}{!}{
\begin{tabular}{>{\centering\arraybackslash}m{2.5cm} p{16cm}}
\toprule
Category & Prompt Template \\
\midrule
Reformat &

Please reformat the provided solution for the given problem by dividing it into multiple detailed steps. 
These steps must explicitly present the final answer within \texttt{\textbackslash boxed\{\}}. 
For each step, enrich the content with the minimal necessary details to enhance clarity. 
Ensure that any added information is precise and unambiguous to avoid potential misunderstandings. 
Return the response in explicit JSON format as follows:

$
\begin{array}{l}
[ \\
\quad \text{"[STEP 1 CONTENT]"}, \\
\quad \text{"[STEP 2 CONTENT]"}, \\
\quad // \text{Continue for each step...}\\
]\\
\end{array}
$ 
\\
\midrule
Post-process & 
Please check and fix any LaTeX formatting errors in the following mathematical solution step. Return only the corrected step with proper LaTeX syntax. \\

\bottomrule
\end{tabular}
}
\label{tab:prompt_template_step_solution}
\end{table*}

\begin{table*}[ht]
\center
\caption{Detailed Results on \textsc{HS-STaR} Invariants.}
\resizebox{1\linewidth}{!}{
\begin{tabular}{cccccccccc}
\toprule
\multicolumn{1}{c}{\textbf{Estimation Strategy}} & \multicolumn{1}{c}{\textbf{Iteration}} & \textbf{GSM8K} & \textbf{\begin{tabular}[c]{@{}c@{}}MATH\\ 500\end{tabular}} & \textbf{\begin{tabular}[c]{@{}c@{}}Olympiad\\Bench \end{tabular}} & \textbf{\begin{tabular}[c]{@{}c@{}}Minerva\\ Math\end{tabular}} & \textbf{\begin{tabular}[c]{@{}c@{}}AMC23 \end{tabular}} & \textbf{\begin{tabular}[c]{@{}c@{}} College\\Math\end{tabular}} & \textbf{AIME24} & \textbf{Avg.} \\ 
\toprule
\multirow{3}{*}{\textsc{HS-STaR}-SDE~(Oracle)} & 1 & 88.8 & 71.8 & 34.4 & 28.3 & 44.8 & 45.2 & 7.3 & 45.8\\
& 2 & 90.2 & 71.6 & 37.2 & 30.9 & 44.2 & 47.2 & 8.3 & 47.1\\
& 3 & 90.6 & 73.8 & 34.5 & 32.7 & 46.2 & 46.7 & 10.9 & 47.9\\
\midrule
\multirow{3}{*}{\textsc{STaR}-DPO} & 1 & 87.4 & 69.8 & 30.8 & 25.4 & 43.7 & 45.4 & 6.7 & 44.2\\
 & 2 & 87.3 & 68.4 & 32.4 & 29.8 & 44.1 & 45.3 & 8.3 & 45.1\\
& 3 & 88.6 & 69.8 & 33.3 & 29.8 & 44.3 & 45.7 & 8.3 & 45.7\\
\midrule
\multirow{3}{*}{\textsc{HS-STaR}-Acc} & 1 & 87.2 & 70.4 & 31.7 & 27.9 & 42.7 & 46.0 & 7.0 & 44.7\\
 & 2 & 88.6 & 71.8 & 32.3 & 32.0 & 45.1 & 45.7 & 8.2 & 46.2\\
& 3 & 89.8 & 71.0 & 35.4 & 29.8 & 46.9 & 46.5 & 8.3 & 46.8\\
\midrule
\multirow{3}{*}{\textsc{HS-STaR}-Reward} & 1 & 87.6 & 70.4 & 33.5 & 28.3 & 45.1 & 45.3 & 7.6 & 45.4\\
& 2 & 89.0 & 70.2 & 33.6 & 31.2 & 45.2 & 46.0 & 8.9 & 46.3\\
& 3 & 89.3 & 73.8 & 35.4 & 28.3 & 44.5 & 46.8 & 9.0 & 46.7\\
\midrule
\rowcolor{ourscolor} & 1 & 88.0 & 69.8 & 33.3 & 29.8 & 45.2 & 46.0 & 7.3 & 45.6 \\
\rowcolor{ourscolor} \textsc{HS-STaR}~(Ours) & 2 & 89.5 & 71.8 & 34.2 & 31.2 & 45.7 & 46.0 & 7.8 & 46.6 \\
\rowcolor{ourscolor} & 3 & 90.3 & 72.8 & 35.9 & 31.6 & 46.5 & 46.4 & 8.9 & 47.5 \\
\bottomrule
\end{tabular}
}
\label{tab:detailed-ablation-on-estimation}
\end{table*}

\subsubsection{\textsc{HS-STaR} Variants}
\noindent\textbf{\textsc{STaR}-DPO.}
This baseline configuration employs standard sampling techniques followed by iterative preference optimization.

\noindent\textbf{\textsc{HS-STaR}-Acc.}
A variant of \textsc{HS-STaR}. In the Difficulty Estimation phase, the estimation is solely based on the accuracy of responses sampled during Pre-Sampling. Subsequently, Re-Sampling is performed on the identified boundary examples. Finally, the collected data from both phases is utilized for preference optimization.

\begin{table*}[ht]
\centering
\caption{Detailed Ablation Study on Re-sampling Strategies.}
\resizebox{1\linewidth}{!}{
\begin{tabular}{cccccccccc}
\toprule
\multicolumn{1}{c}{\textbf{Re-Sampling Strategy}} & \multicolumn{1}{c}{\textbf{Iteration}} & \textbf{GSM8K} & \textbf{\begin{tabular}[c]{@{}c@{}}MATH\\ 500\end{tabular}} & \textbf{\begin{tabular}[c]{@{}c@{}}Olympiad\\Bench \end{tabular}} & \textbf{\begin{tabular}[c]{@{}c@{}}Minerva\\ Math\end{tabular}} & \textbf{\begin{tabular}[c]{@{}c@{}}AMC23 \end{tabular}} & \textbf{\begin{tabular}[c]{@{}c@{}} College\\Math\end{tabular}} & \textbf{AIME24} & \textbf{Avg.} \\ 
\midrule
\multirow{3}{*}{w/o Re-Sampling} & 1 & 87.4 & 69.8 & 30.8 & 25.4 & 43.7 & 45.4 & 6.7 & 44.2 \\
& 2 & 87.3 & 68.4 & 32.4 & 29.8 & 44.1 & 45.3 & 8.3 & 45.1 \\
& 3 & 88.6 & 69.8 & 33.3 & 29.8 & 44.3 & 45.7 & 8.3 & 45.7 \\
\midrule
\multirow{3}{*}{Inlier}  & 1 & 86.7 & 65.8 & 30.5 & 23.2 & 39.0 & 43.4 & 5.7 & 42.0 \\
& 2 & 87.3 & 67.4 & 30.8 & 24.6 & 40.3 & 44.4 & 6.1 & 43.0 \\
& 3 & 87.2 & 70.0 & 32.1 & 27.6 & 41.2 & 45.2 & 6.2 & 44.2 \\
\midrule
\multirow{3}{*}{Outlier} & 1 & 85.7 & 66.2 & 29.8 & 25.4 & 36.1 & 42.7 & 6.2 & 41.7 \\
& 2 & 86.4 & 67.4 & 29.6 & 25.7 & 37.3 & 42.2 & 5.0 & 41.9 \\
& 3 & 86.1 & 67.2 & 30.4 & 25.0 & 37.7 & 43.2 & 5.7 & 42.2 \\
\midrule
\multirow{3}{*}{Inlier + Boundary} & 1 & 87.7 & 69.8 & 33.8 & 27.6 & 44.9 & 45.3 & 8.1 & 45.3 \\
& 2 & 89.2 & 71.6 & 35.1 & 29.4 & 44.9 & 45.7 & 8.8 & 46.4 \\
& 3 & 89.5 & 72.0 & 36.1 & 33.1 & 45.0 & 46.5 & 8.5 & 47.2 \\
\midrule
\multirow{3}{*}{Boundary + Outlier} & 1 & 87.1 & 70.6 & 32.0 & 24.6 & 42.5 & 45.2 & 7.2 & 44.2 \\
& 2 & 87.1 & 69.6 & 32.1 & 27.6 & 43.2 & 45.5 & 8.1 & 44.7 \\
& 3 & 88.4 & 71.2 & 33.6 & 30.5 & 44.6 & 45.7 & 8.2 & 46.0 \\
\midrule
\multirow{3}{*}{Inlier + Outlier} & 1 & 86.8 & 65.8 & 30.2 & 24.6 & 38.5 & 42.8 & 4.9 & 41.9 \\
& 2 & 86.0 & 66.8 & 28.9 & 23.5 & 37.3 & 43.3 & 4.8 & 41.5 \\
& 3 & 86.1 & 67.8 & 30.4 & 25.4 & 40.4 & 43.7 & 6.1 & 42.8 \\
\midrule
\rowcolor{ourscolor} & 1 & 88.0 & 69.8 & 33.3 & 29.8 & 45.2 & 46.0 & 7.3 & 45.6 \\
\rowcolor{ourscolor} Boundary & 2 & 89.5 & 71.8 & 34.2 & 31.2 & 45.7 & 46.0 & 7.8 & 46.6 \\
\rowcolor{ourscolor} & 3 & 90.3 & 72.8 & 35.9 & 31.6 & 46.5 & 46.4 & 8.9 & 47.5 \\
\bottomrule
\end{tabular}
}
\label{tab:detailed-ablation-on-samples-selection}
\end{table*}

\noindent\textbf{\textsc{HS-STaR}-Reward}
A variant of \textsc{HS-STaR}. In the Difficulty Estimation phase, the estimation is solely based on the reward of responses sampled during Pre-Sampling. Subsequently, Re-Sampling is performed on the identified boundary examples. Finally, the collected data from both phases is utilized for preference optimization.

\noindent\textbf{\textsc{HS-STaR}-SDE}
A variant of \textsc{HS-STaR}. In the Difficulty Estimation phase, the estimation is based on SDE method. Subsequently, Re-Sampling with the full sampling budget is conducted on the identified boundary examples. Finally, the collected data from both phases is utilized for preference optimization.

\subsubsection{Re-Sampling Strategies}

\noindent\textbf{w/o Re-Sampling.} This configuration serves as a standard baseline, employing conventional sampling techniques followed by iterative preference optimization without difficulty-based re-sampling.

\noindent\textbf{Re-Sampling on Inlier.} Following prior difficulty estimation in the pre-sampling phase, remaining sampling efforts are exclusively focused on Inlier samples for subsequent iterative preference optimization.

\noindent\textbf{Re-Sampling on Outlier.} Following prior difficulty estimation in the pre-sampling phase, remaining sampling efforts are exclusively focused on Outlier samples for subsequent iterative preference optimization.

\noindent\textbf{Re-Sampling on Inlier + Outlier.} Following prior difficulty estimation in the pre-sampling phase, remaining sampling efforts are allocated to both Inlier and Outlier samples for subsequent iterative preference optimization.

\noindent\textbf{Re-Sampling on Inlier + Boundary.} Following prior difficulty estimation in the pre-sampling phase, remaining sampling efforts are allocated to both Inlier and Boundary samples for subsequent iterative preference optimization.

\noindent\textbf{Re-Sampling on Outlier + Boundary.} Following prior difficulty estimation in the pre-sampling phase, remaining sampling efforts are allocated to both Outlier and Boundary samples for subsequent iterative preference optimization.

\noindent\textbf{Re-Sampling on Boundary (Ours).} Following prior difficulty estimation in the pre-sampling phase, remaining sampling efforts are exclusively focused on Boundary samples for subsequent iterative preference optimization.

\section{Algorithm}

The overall procedure of our algorithm is illustrated in Algorithm~\ref{alg:main_framework}.

\section{Prompt Template}

To construct the stepwise warmup dataset, we leveraged the MATH dataset (Hendrycks et al., 2021) and prompted GPT-4o-2024-08-06 to systematically rewrite each solution in a JSON format. Subsequently, these rewritten solutions were separated by the delimiter ``\verb|\n\n|''. The prompt template used for this process is presented in Table \ref{tab:prompt_template_step_solution}. We initially employed a Reformat prompt to guide the model in restructuring the solutions in json format. In cases where the Reformat attempt failed, a Post-process prompt was utilized to further refine or reshape the output. Finally, the resulting data was filtered based on the provided answer.

\section{Additional Experimental Results}

\subsection{Iterative Results on Qwen2.5-7B}
\begin{figure}[t]
\centering
\includegraphics[width=\linewidth]{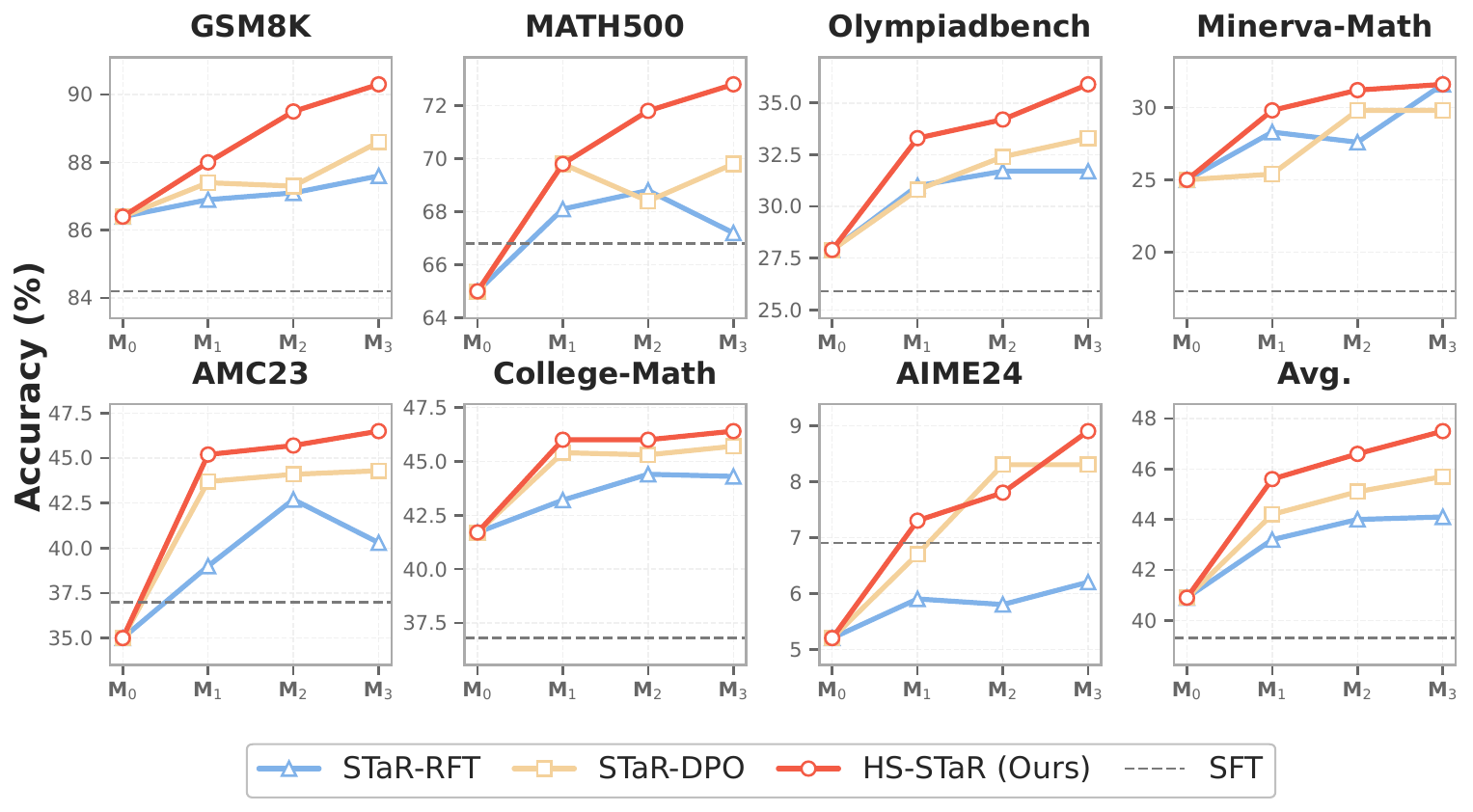}
\caption{
Comparison of the performance improvements of Qwen2.5-7B across three training iterations.
}
\label{fig:iterative_qwen2.5_7b}
\end{figure}

As illustrated in Fig.~\ref{fig:iterative_qwen2.5_7b}, \textsc{HS-STaR} consistently outperformed all baseline methods across all evaluated benchmarks. 
As the number of iterations increased, the performance of all methods gradually improved; notably, \textsc{HS-STaR} attained the highest $M_3$ accuracy on every benchmark.
Moreover, the overall average accuracy highlights that \textsc{HS-STaR} delivers the most substantial improvement compared to other approaches.

\subsection{Results on \textsc{HS-STaR} Invariants}
As shown in Table~\ref{tab:detailed-ablation-on-estimation}, we additionally present the performance of various Difficulty Estimation ablation strategies across different evaluation datasets at each iterative round.

\subsection{Results on Re-sampling Strategies}
As shown in Table~\ref{tab:detailed-ablation-on-samples-selection}, we also provide the performance of various difficulty Re-Sampling ablation strategies across different evaluation datasets at each iteration round.

\end{document}